\documentclass[authoryear,preprint,review,12pt]{elsarticle}
\usepackage{lineno}
\modulolinenumbers[5]
\usepackage{amssymb}
\usepackage{epsfig}
\usepackage{subfigure}
\usepackage{multirow}
\usepackage{float}
\usepackage{amsmath}
\usepackage{longtable}
\usepackage{booktabs}
\usepackage{color}
\usepackage{subfigure}
\usepackage{fancyhdr}
\usepackage{epstopdf}
\usepackage{hyperref}
\usepackage{algorithm}
\usepackage{algorithmic}
\usepackage{ulem}
\usepackage{url}
\usepackage{bm}
\usepackage{amsopn}
\usepackage{amssymb,amsmath,color,times}
\usepackage{graphicx}
\usepackage{latexsym}
\usepackage{amsfonts}
\usepackage{enumerate}
\usepackage{tabularx}
\usepackage{ifpdf}
\usepackage{colortbl,booktabs}
\usepackage{xcolor}
\usepackage{pifont}       
\usepackage{bbding}       
\usepackage{fontawesome}

\usepackage{ragged2e}
\justifying

\usepackage{amssymb}
\usepackage{amsmath}

\journal{Nuclear Physics B}

\begin{document}

\begin{frontmatter}



\title{SAM-Based Building Change Detection with Distribution-Aware Fourier Adaptation and Edge-Constrained Warping} 

\author[mymainaddress]{Yun-Cheng~Li}

\author[mymainaddress]{Sen~Lei}

\author[mymainaddress]{Yi-Tao~Zhao}

\author[mymainaddress]{Heng-Chao~Li\corref{mycorrespondingauthor}}
\cortext[mycorrespondingauthor]{Corresponding author}
\ead{lihengchao_78@163.com}

\author[mysecondaryaddress]{Jun~Li}

\author[mythirdaryaddress]{Antonio~Plaza}

\address[mymainaddress]{School of Information Science and Technology, Southwest Jiaotong University, Chengdu 611756, China.}

\address[mysecondaryaddress]{School of Computer Science, China University of Geosciences, Wuhan 430074, China.}
\address[mythirdaryaddress]{Department of Technology of Computers and Communications, University of Extremadura,Caceres 10003, Spain.}
\begin{abstract}
Building change detection remains challenging for urban development, disaster assessment, and military reconnaissance. While foundation models like Segment Anything Model (SAM) show strong segmentation capabilities, SAM is limited in the task of building change detection due to domain gap issues. Existing adapter-based fine-tuning approaches face challenges with imbalanced building distribution, resulting in poor detection of subtle changes and inaccurate edge extraction. Additionally, bi-temporal misalignment in change detection, typically addressed by optical flow, remains vulnerable to background noises. This affects the detection of building changes and compromises both detection accuracy and edge recognition. To tackle these challenges, we propose a new SAM-Based Network with Distribution-Aware Fourier Adaptation and Edge-Constrained Warping (FAEWNet) for building change detection. FAEWNet utilizes the SAM encoder to extract rich visual features from remote sensing images. To guide SAM in focusing on specific ground objects in remote sensing scenes, we propose a Distribution-Aware Fourier Aggregated Adapter to aggregate task-oriented changed information. This adapter not only effectively addresses the domain gap issue, but also pays attention to the distribution of changed buildings. Furthermore, to mitigate noise interference and misalignment in height offset estimation, we design a novel flow module that refines building edge extraction and enhances the perception of changed buildings. Our state-of-the-art results on the LEVIR-CD, S2Looking and WHU-CD datasets highlight the effectiveness of FAEWNet. The code is available at https://github.com/SUPERMAN123000/FAEWNet.
\end{abstract}

\begin{keyword}
Remote sensing, building change detection, foundation model, segment anything model (SAM).
\end{keyword}

\end{frontmatter}

\section{Introduction}
With the advancement of remote sensing technology for earth observation, remote sensing image change detection has become a crucial research area. Building change detection focuses on monitoring the dynamic attributes of remote sensing surfaces over time, which are influenced by natural elements and human activities. Accurate detection of these changes is a powerful tool for applications such as disaster assessment (\citeauthor{hansch2022}, \citeyear{hansch2022}), and urban expansion (\citeauthor{Ren2024}, \citeyear{Ren2024}). 

Traditional change detection methods (\citeauthor{cao2014}, \citeyear{cao2014}; \citeauthor{bovolo2008}, \citeyear{bovolo2008}) primarily classify individual pixels using thresholds to generate binary change maps. Their performance in complex scenarios is often unsatisfactory due to their reliance on decision classifiers and hand-crafted thresholds.

In recent years, deep learning has significantly advanced various fields (\citeauthor{Lei2024}, \citeyear{Lei2024}; \citeauthor{like2023}, \citeyear{like2023}; \citeauthor{TF2024}, \citeyear{TF2024}), especially in building change detection (\citeauthor{Pang2021}, \citeyear{Pang2021}; \citeauthor{Zhao2024}, \citeyear{Zhao2024}; \citeauthor{Pang2024}, \citeyear{Pang2024}; \citeauthor{lky2025}, \citeyear{lky2025}; \citeauthor{like2024}, \citeyear{like2024}). Convolutional neural network (CNN)-based algorithms (\citeauthor{Codegoni2022}, \citeyear{Codegoni2022}; \citeauthor{CH2022}, \citeyear{CH2022}; \citeauthor{LY2021}, \citeyear{LY2021}; \citeauthor{ZYT2024}, \citeyear{ZYT2024}) effectively extract robust features from the changed areas, achieving strong performance in complex scenarios. Transformer-based methods (\citeauthor{Bandara2022}, \citeyear{Bandara2022}; \citeauthor{XC2024}, \citeyear{XC2024}; \citeauthor{FJF2024}, \citeyear{FJF2024}) have further accelerated developments by capturing long-range dependencies across entire images, providing models with global receptive fields. This shift has opened new avenues for building change detection, which requires high-level semantic understanding. Despite their notable success, these methods are still far from being practically applicable in complex and rapidly evolving spatial-temporal environments. Moreover, as the scale of the model increases, the limited availability of annotated data for building change detection significantly limits the potential of these methods. 

\begin{figure}[!htb]
\centering
\includegraphics[width=1\linewidth]{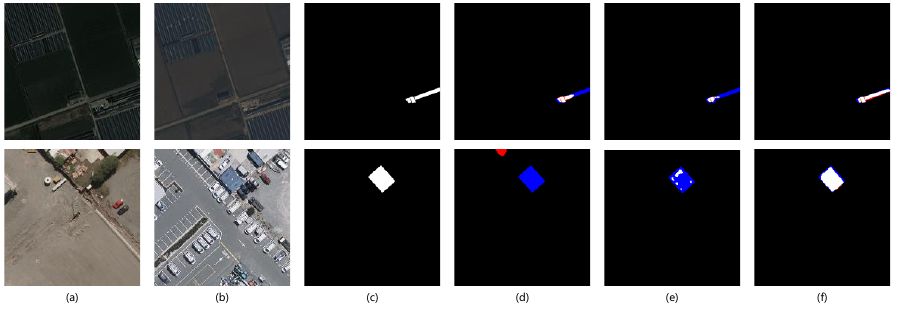}
\vspace{-0.4cm}
\caption{Cases of Existing Foundation Models with Imbalanced Building Distribution. (a) T1 image, (b) T2 image, (c) Change label, (d) BAN (\citeauthor{LiKaiyu2024}, \citeyear{LiKaiyu2024}), (e) TTP (\citeauthor{Chen2023}, \citeyear{Chen2023}), (f) FAEWNet. White represents a true positive, black is a true negative, red indicates a false positive, and blue stands as a false negative.}
\label{fig:shili}
\end{figure}

Recently, visual foundation models have been widely applied in computer vision. Such foundation models leverage large-scale pretraining to capture transferable visual representations, such as Contrastive Language-Image Pre-Training (CLIP) methods (\citeauthor{Radford2021}, \citeyear{Radford2021}) and Segment Anything Model (SAM) (\citeauthor{kirillov2023}, \citeyear{kirillov2023}). The bi-temporal adapter network (BAN) (\citeauthor{LiKaiyu2024}, \citeyear{LiKaiyu2024}) which was based on CLIP, integrated foundation model knowledge into change detection through a dual-temporal adapter network. Furthermore, SAM and its variants (e.g., FastSAM (\citeauthor{Zhao2023}, \citeyear{Zhao2023}), Mobile SAM (\citeauthor{Zhang2023}, \citeyear{Zhang2023})), have emerged as promising solutions to challenges in building change detection. \citeauthor{ZD2024} (\citeyear{ZD2024}) integrated SAM with feature interaction for remote sensing change detection in their SFCDNet. \citeauthor{WM2024} (\citeyear{WM2024}) proposed weakly supervised change detection based on multiscale class activation map (CAM) fusion and efficient SAM fine-tuning. \citeauthor{Chen2023} (\citeyear{Chen2023}) fine-tuned SAM using LoRA and employed time-traveling activation gates in the encoder to facilitate interaction between temporal phases. However, these models face limitations in building change detection, primarily due to inductive biases learned from natural images, resulting in suboptimal performance in remote sensing images. While existing adapter fine-tuning methods have made progress in bridging the domain gap between natural and remote sensing images, they overlook the imbalance in building distribution. This oversight hinders SAM's ability to detect subtle building changes and extract building edges effectively, as shown in Figure \ref{fig:shili}. Moreover, existing optical flow methods struggle to detect building height offsets due to background noises, leading to reduced precision and difficulties in edge recognition.

In this paper, we propose a Distribution-Aware Fourier Aggregated Adapter to fine-tune SAM's robust semantic exploitation capabilities, enabling better handling of the imbalanced distribution of buildings in remote sensing images. By incorporating this novel adapter, we enhance SAM's ability to focus on subtle building changes and improve its edge extraction performance, particularly in regions with weak contrast or small-scale changes. To address inaccurate building edge extraction caused by background noises in existing optical flow modules during building height offset detection, we propose a novel optical flow module. The main contributions of this paper are summarized as follows:
\begin{itemize}
\item We propose a novel method, FAEWNet, for building change detection. Unlike previous methods, FAEWNet utilizes a distribution-aware Adapter and global alignment to effectively extract buildings that have undergone subtle and complete changes.

\item We propose a novel fine-tuning strategy, the Distribution-Aware Fourier Aggregated (DAFA) Adapter, which captures robust semantic changes from bitemporal image features to perceive the distribution of buildings, enabling better extraction of subtle buildings and edge details.

\item We design an interactive fusion module, the Multiscale Aware Flow Aggregation (MSAFA) module, which addresses misalignment in bitemporal images from both local and global perspectives. Compared to previous work, it helps hinder background noises. 

\end{itemize}

\section{Related Work}
\subsection{Building Change Detection}
Accurate building change detection is crucial to analyze remote sensing images. In recent years, various change detection approaches have been proposed, with CNN-based methods achieving notable success in extracting robust features from changed buildings. \cite{Daudt2018} introduced three U-Net-based fully convolutional Siamese networks: FC-EF, FC-Siam-conc, and FC-Siam-diff. Building upon the FC-Siam-conc model, \citeauthor{Zhang2020b} (\citeyear{Zhang2020b}) incorporated an attention mechanism to develop stronger basic blocks and added deep supervision, leading to faster convergence and improved performance. \citeauthor{Song2021} (\citeyear{Song2021}) utilized spatial attention to distinguish between changed and background pixels of a building, combining channel attention with an \textit{atrous} spatial pyramid pooling module to enhance multiscale contextual information. 

The emergence of Vision Transformer (ViT) has paved new pathways for change detection. \citeauthor{Wang2021} (\citeyear{Wang2021}) explored the potential of Transformers in change detection, proposing a Transformer-based network. \citeauthor{Zhou2022} (\citeyear{Zhou2022}) utilized the Multi-Head Self-Attention (MHSA) mechanism, a key component of Transformers, to build their change detection network. \cite{Chen2021a} used a Transformer encoder to identify the changed area of interest, while two Siamese Transformer decoders refined the resulting change maps. ChangeFormer (\citeauthor{Bandara2022}, \citeyear{Bandara2022}) inspired by Segformer (\citeauthor{Xie2021}, \citeyear{Xie2021}), was a Transformer-based change detection model. Its basic module applied spatial downsampling of the query, key, and value within the self-attention mechanism to reduce computational complexity. Currently, research on foundation models has become a mainstream topic. In this study, we evaluate the performance of SAM in building change detection.

\subsection{SAM for Building Change Detection}
A major challenge in applying deep neural networks to building change detection is their dependence on large, well-annotated training dataset. To overcome this challenge, researchers have explored pretraining vision models on web-scale datasets to develop universal recognition capabilities for building change detection. A notable example is CLIP, trained on 400 million image-text pairs, which enables it to describe visual content using text. Denoising diffusion probabilistic models (\citeauthor{Bandara2022Ddpm-cd}, \citeyear{Bandara2022Ddpm-cd}) were pretrained on large-scale remote sensing images and then used as feature extractors for building change detection. In ChangeCLIP (\citeauthor{DONG2024}, \citeyear{DONG2024}), semantic information from image-text pairs was used to extract bitemporal features and capture detailed semantic changes with a differential features compensation module. 

SAM is a segmentation model trained on millions of annotated images, enabling zero-shot generalization to unseen images and objects. SAM has been widely utilized in various Remote Sensing fields (\citeauthor{LNQ2025}, \citeyear{LNQ2025}; \citeauthor{YZY2023}, \citeyear{YZY2023}). SAM has also been applied to change detection tasks. For instance, SAM-CD (\citeauthor{Ding2024}, \citeyear{Ding2024}) used Fast-SAM as a static visual encoder and fine-tuned adapter network with the change decoder in a fully supervised manner on change detection datasets. In  Segment Anything Model-UNet Change Detection Model (\citeauthor{ZXQ2024}, \citeyear{ZXQ2024}), SAM was embedded into the model to enhance attention to key information and improve feature extraction capabilities. \citeauthor{Sun2024} (\citeyear{Sun2024}) proposed a framework that utilized SAM’s semantic knowledge with multiscale feature extraction and masked attention for accurate change detection. In this paper, we propose an innovative fine-tuning strategy: the Distribution-Aware Fourier Aggregation (DAFA) Adapter, which enhances the model's adaptability and detection performance in complex remote sensing image scenarios.
\begin{figure*}[t]
  \centering
  {\includegraphics[width=1\linewidth]{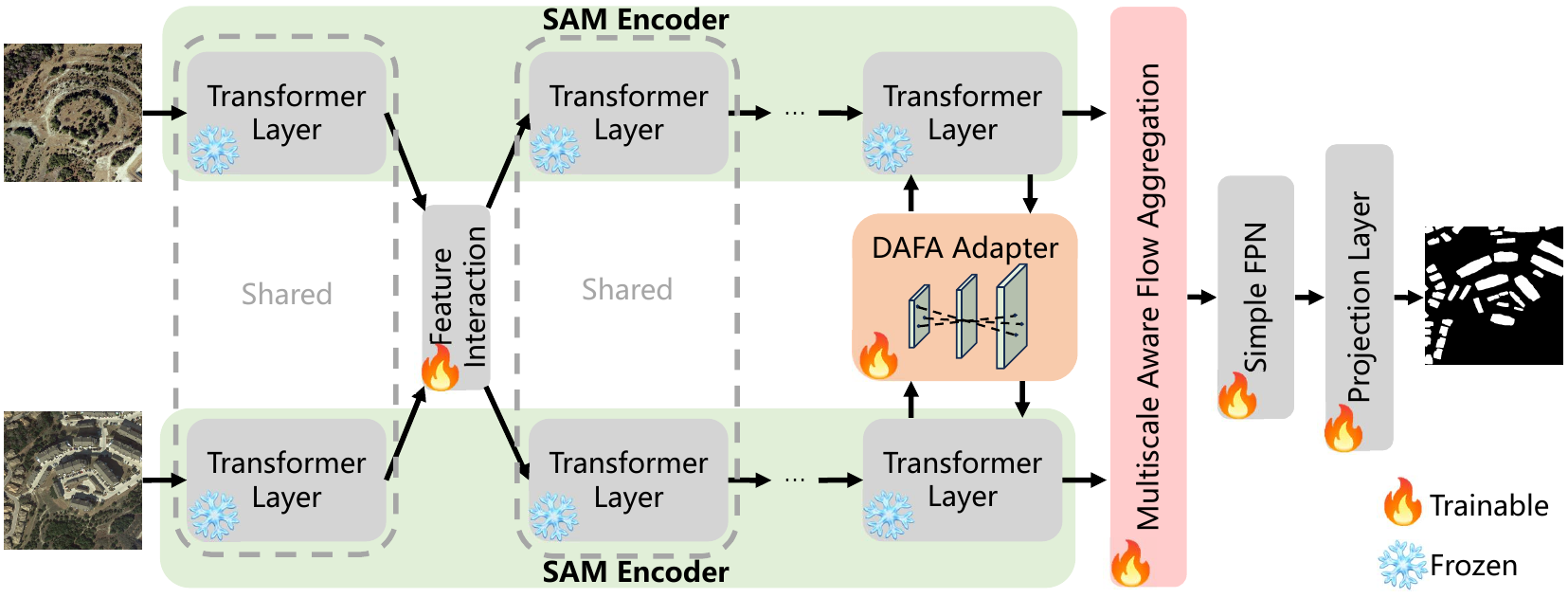}}
  \caption{The overall framework of FAEWNet. The backbone used is SAM. FAEWNet takes as input a pair of features from very high-resolution remote sensing images.}
  \label{fig:overallframework}
\end{figure*}
\section{Methodology}
\subsection{Overall Architecture}
To address the significant annotation requirements in building change detection, we leverage the broad knowledge provided by foundation model. In this work, we propose a change detection network termed FAEWNet, built upon the segmentation features of SAM. As illustrated in Figure \ref{fig:overallframework}, FAEWNet is a weight-sharing Siamese network with three key components: an encoder, a head, and a decoder. Given paired input images $\mathcal{I}_0$ and $\mathcal{I}_1$, the resulting building change map $\mathcal{Y}$ can be represented as:

\begin{equation}
    \label{eqn:1}
    \mathcal{Y} = \operatorname{Dec}(\operatorname{Head}(\operatorname{Enc}(\mathcal{I}_{0}), \operatorname{Enc}(\mathcal{I}_{1})))
\end{equation}
where $\operatorname{Enc}(\cdot)$, $\operatorname{Head}(\cdot)$, and $\operatorname{Dec}(\cdot)$ denote the encoder, fusion head, and decoder (which incorporates FPN and projection layers), respectively. 

The FAEWNet encoder performs hierarchical processing in two stages. Let $\mathcal{F}_{i,j}$ denote the output at a specific hierarchical stage within the encoder, where $i$ represents the hierarchy level, and $j$ represents the temporal dimension of input images ($i \in \{0, 1, 2, ..., 23\}$, and $j \in \{0, 1\}$). Firstly, the multilevel features $\mathcal{F}_{i,j}$ are processed by a shared-weight SAM stage, yielding modified features $\mathcal{F}^{\prime}_{i,j}$. Next, the modified features $\mathcal{F}^{\prime}_{i,j}$ for all temporal instances $\mathcal{I}_{j}$ are fed into Time-traveling Activation Gate (TTAG) (\citeauthor{Chen2023}, \citeyear{Chen2023}) modules to yield correlated features $\mathcal{F}_{i+1,j}$. 

In contrast to \cite{Chen2023}, which applies TTAG after global attention, we place it after local attention instead. In the final stage, we integrate our proposed the DAFA Adapter into the Transformer layers to address the domain gap between remote sensing and natural images. The features for each hierarchy $i$ are as follows: 
\begin{equation}
    \label{eqn:2}
    \begin{aligned}
    &\mathcal{F}^{\prime}_{i+1,j}= \operatorname{Stage}_{i}(\mathcal{F}_{i,j}) ,\quad \forall j \\
    &\mathcal{F}_{i+1,j} = \operatorname{TTAG}(\mathcal{F}^{\prime}_{i+1,0},\mathcal{F}^{\prime}_{i+1,1}) , i \neq 5, 11, 17, 23,\quad \forall j\\
    &\mathcal{F}_{i+1,j} = \operatorname{DAFA}(\mathcal{F}^{\prime}_{i+1,0},\mathcal{F}^{\prime}_{i+1,1}) , i = 23,\quad \forall j
    \end{aligned}
\end{equation}
where $\operatorname{TTAG}(\cdot)$ denotes the Time-traveling Activation Gate and $\operatorname{DAFA}(\cdot)$ represents the Distribution-Aware Fourier Aggregation Adapter.

To effectively address building offsets from multiscale and comprehensive perspectives, we utilize MSAFA module to fuse the two feature maps $\mathcal{F}_0$ and $\mathcal{F}_1$, produced by the encoder. These feature maps are then fed into the network's decoder to generate the final change map $\mathcal{Y} \in \mathbb{R}^{2 \times H \times W}$, formulated as:

\begin{equation}
    \label{eqn:3}
    \begin{aligned}
    \mathcal{Y} &= \operatorname{Project}(\operatorname{FPN}((\operatorname{Fuse}({\mathcal{F}_0},\mathcal{F}_1)))\\
    \end{aligned}
\end{equation}
where $\operatorname{Fuse}(\cdot)$ denotes the proposed MSAFA module, designed to fuse $\mathcal{F}_0$ and $\mathcal{F}_1$. The operation $\operatorname{FPN}(\cdot)$ represents a simple feature pyramid module. The operation $\operatorname{Project}(\cdot)$ represents a projection layer composed of the concatenation of feature maps after $\operatorname{FPN}$ and two $1 \times 1$ convolutional layers.

\subsection{Distribution-Aware Fourier Aggregated Adapter}
\begin{figure}[!htp]
\centering
{\includegraphics[width=1\linewidth]{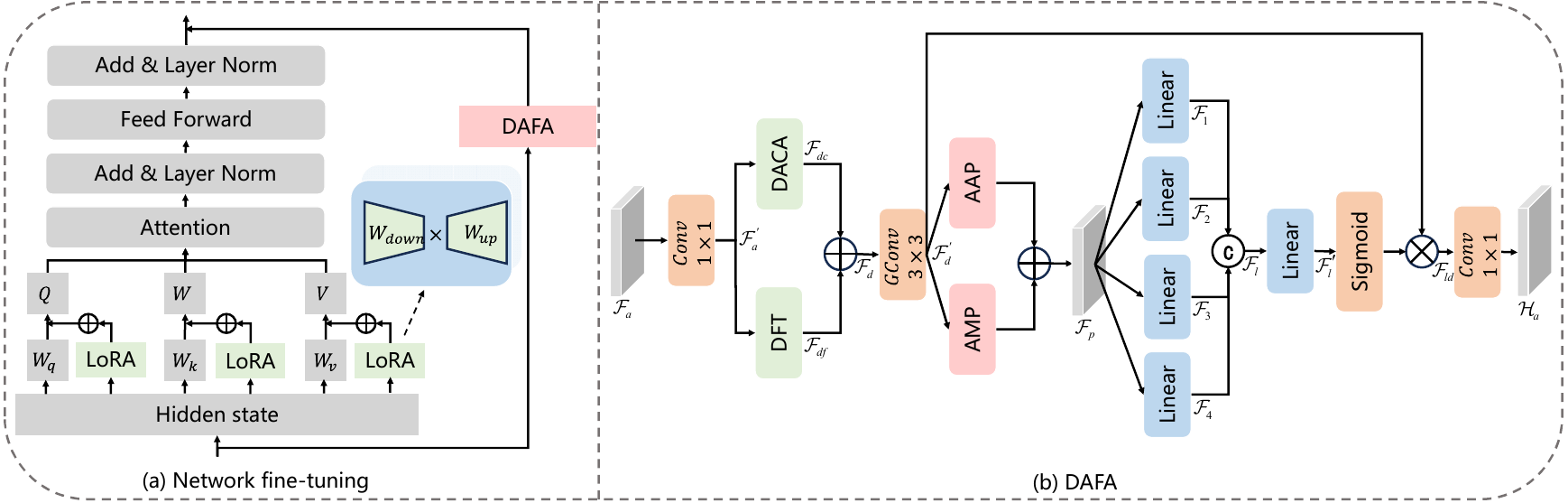}}
\caption{(a) Adapter for fine-tuning network, (b) DAFA for fine-tuning network.}
\label{fig:Adapter_DAFA}
\end{figure}

While SAM demonstrates remarkable capabilities in semantic representation learning for natural images, its direct application to building change detection in remote sensing images faces critical limitations. Optical images and remote sensing data exhibit fundamental differences in feature distributions – buildings often manifest as small, densely distributed, and spectrally similar objects with subtle structural variations. The SAM architecture, utilizing four global attention layers among 24 Transformer blocks, insufficiently captures these fine-grained distribution patterns, particularly for building edges and subtle building changes. To enhance the model's sensitivity to building distributions, especially for small buildings, we propose a novel DAFA Adapter into SAM after the final global attention layer. DAFA can integrate spectral-domain analysis and distribution-sensitive feature recalibration. In addition, we introduce low-rank trainable parameters into the multihead attention layers, as shown in Figure \ref{fig:Adapter_DAFA} (a). This method enables efficient fine-tuning of the foundation model.

The specific stucture of DAFA is shown in Figure \ref{fig:Adapter_DAFA} (b). First, we pass $\mathcal{F}_{a}$ through a $1\times1$ convolutional module to generate $\mathcal{F}^{\prime}_{a}$. Next, $\mathcal{F}^{\prime}_{a}$ is fed into Distribution-Aware Channel Attention (DACA) (\citeauthor{ZGC2024}, \citeyear{ZGC2024}) module. The channel-wise statistics are collected from $\mathcal{F}^{\prime}_{a}$ to reduce information loss caused by real-valued activation. This process captures the distribution properties of the activation. Inspired by effective channel attention module, a 1D convolution is applied followed by a Sigmoid function to derive a distribution-aware scale factor from these statistics, thereby minimizing computational overhead and parameter requirements. Different from \cite{ZGC2024}, we multiply the distribution scale factor directly by $\mathcal{F}^{\prime}_{a}$ to maintain the characterization of the original buildings. As given by the following equation:
\begin{equation}
    \label{eqn:5}
    \begin{aligned}
    &\mathcal{F}_{da} = \operatorname{Concat}(\operatorname{AAP}(|\mathcal{F}^{\prime}_{a}|),\operatorname{Mean}(\mathcal{F}^{\prime}_{a}),\operatorname{Std}(\mathcal{F}^{\prime}_{a}))\\
    &\mathcal{F}_{dc} = \operatorname{Conv}_{1\times1}(\operatorname{Sigmoid}(\operatorname{Conv1d}(\mathcal{F}_{da}))\cdot\mathcal{F}^{\prime}_{a})
    \end{aligned}
\end{equation}
where $\operatorname{AAP}(\cdot)$ represents the adaptive average pooling operation, $|\cdot|$ is the absolute value, $\operatorname{Mean}(\cdot)$ indicates the mean value of each channel, $\operatorname{Std}(\cdot)$ represents the standard deviation of each channel. Different from \cite{ZGC2024}, $\operatorname{AAP}(|\mathcal{F}^{\prime}_{a}|)$ detects significant structural changes or new/demolished parts of the buildings by capturing the intensity of the changes. $\operatorname{Mean}(\mathcal{F}^{\prime}_{a})$ focuses on global brightness changes in buildings and can help detect large-scale building changes. $\operatorname{Std}(\mathcal{F}^{\prime}_{a})$ emphasizes local changes, particularly changes in building details, facilitating the identification of complex local changes.

 The Discrete Fourier Transform (DFT) layer can detect structural changes that are geometrically subtle but spectrally significant in the frequency domain. Therefore we integrate a DFT layer to enhance the recognition of changes, such as new construction, demolitions or renovations. Specifically, the Fourier transform is first applied along the feature vector dimension, followed by the sequence dimension, with only the real part retained. The decision to preserve only the real part is intentional, as it better aligns with the specific requirements of building change detection. The real part of the Fourier transform contains crucial information about the spatial patterns and structures of the buildings, which is essential for detecting building changes. The formula is as follows:
\begin{equation}
    \label{eqn:6}
    \mathcal{F}_{df}=\mathbb{R}(\operatorname{Seq}(\operatorname{Hidden}(\mathcal{F}^{\prime}_{a}))) 
\end{equation}
where $\operatorname{Seq}(\operatorname{Hidden}(\cdot))$ represents the Discrete Fourier Transform, and $\mathbb{R}(\cdot)$ denotes the real part operation.

The $\mathcal{F}_{dc}$ and $\mathcal{F}_{df}$ are combined to generate $\mathcal{F}_{d}$. $\mathcal{F}_{d}$ is then processed by a $3 \times 3$ group convolution. After this operation, the outputs are passed through adaptive average pooling (AAP) and adaptive max pooling (AMP). This transformation preserves key features, allowing them to pass through the module and enhance the network's representational capacity. To further enhance this, we propose increasing the cardinality of the FC layer in this operation. By integrating multi-branch dense layers with reduction, the model learns a multi-stage channel-wise feature recalibration without spatial information loss. Specifically, the $\operatorname{Linear}$ combinations across feature channels, projecting high-dimensional representations into compact manifolds that emphasize building-specific characteristics. The formula is described in the following:

\begin{equation}
    \label{eqn:7}
    \begin{aligned}
    &\mathcal{F}_{d} = \mathcal{F}_{dc} + \mathcal{F}_{df}\\
    &\mathcal{F}^{\prime}_{d} = \operatorname{GConv_{3 \times 3}}(\mathcal{F}_{d} )\\
    &\mathcal{F}_{p}= \operatorname{AAP}(\mathcal{F}^{\prime}_{d} )+\operatorname{AMP}(\mathcal{F}^{\prime}_{d} )\\
    &\mathcal{F}_{i}= \operatorname{Linear}(\mathcal{F}_{p}), 1\leq i \leq4
    \end{aligned}
\end{equation}
where $\operatorname{GConv}_{3 \times 3}(\cdot)$ denotes $3 \times3$ group convolution, $\operatorname{AAP}(\cdot)$ represents the adaptive average pooling operation, and $\operatorname{AMP}(\cdot)$ refers to the adaptive max pooling operation.

We then concatenate $\mathcal{F}_i (1\leq i\leq 4)$ to generate $\mathcal{F}_{l}$. Next, a $\operatorname{Linear}$ and $\operatorname{Sigmoid}$ are applied to $\mathcal{F}_{l}$, sequentially. The resulting feature map is multiplied by $\mathcal{F}^{\prime}_{d}$ to yield $\mathcal{F}_{ld}$. This approach facilitates the construction of deeper networks while mitigating vanishing gradient issues. Finally, $\mathcal{H}_{a}$ is obtained by applying a $1\times1$ convolutional module to $\mathcal{F}_{ld}$. Here is the formula:  
\begin{equation}
    \label{eqn:8}
    \begin{aligned}
    &\mathcal{F}_{l} = \operatorname{Concat}(\mathcal{F}_i), 1\leq i \leq4\\
    &\mathcal{F}_{ld} = \operatorname{Sigmoid}(\operatorname{Linear}(\mathcal{F}_{l})) \times \mathcal{F}^{\prime}_{d}\\
    &\mathcal{H}_{a} = \operatorname{Conv_{1 \times 1}}(\mathcal{F}_{ld})
    \end{aligned}
\end{equation}
in which $\operatorname{Concat}(\cdot)$ represents the operation of concatenation. 

The DAFA module effectively bridges the gap between natural and remote sensing images by integrating building change information into SAM, while preserving the original architecture of the network.

\subsection{Multiscale Aware Flow Aggregation Module}
\begin{figure}[!htp]
\centering
{\includegraphics[width=1\linewidth]{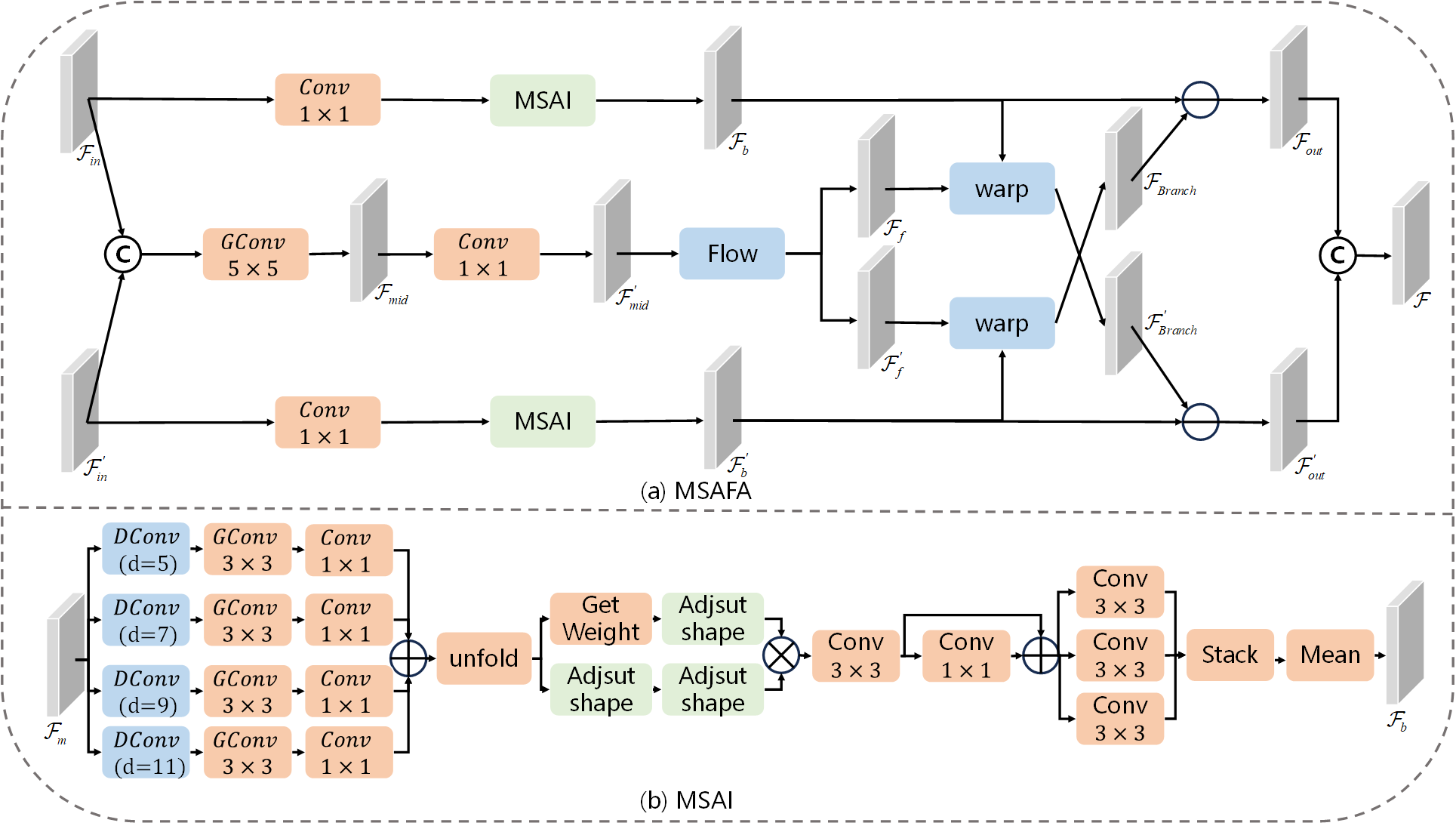}}
\caption{(a) The architecture of MSAFA module. (b) The architecture of MSAI module.}
\label{fig:MSAFA}
\end{figure}
In the field of building change detection, registration errors are a common challenge, with image misalignment often leading to issues such as side-looking distortions. While existing methods (\citeauthor{LMX2022}, \citeyear{LMX2022}; \citeauthor{Li2024}, \citeyear{Li2024}) effectively mitigate these issues using optical flow, they often overlook the global context information, which can hinder accurate building change detection. Unlike previous approaches, our MSAFA module captures multiscale context information to tackle above issues.

The structure of MSAFA module is shown in Figure \ref{fig:MSAFA}, where $\mathcal{F}_{in}$ and $\mathcal{F}^{\prime}_{in}$ are concatenated for further computation. $\mathcal{F}_{mid}$ is generated by passing the concatenated representation through a $5\times5$ group convolutional module. This operation improves the network's ability to extract information about building changes. The equation for $\mathcal{F}_{mid}$ is given as below:
\begin{equation}
    \label{eqn:9}
    \mathcal{F}_{mid} = \operatorname{GConv}_{5\times5}(\operatorname{Concat}(\mathcal{F}_{in},\mathcal{F}^{\prime}_{in}))
\end{equation}
where $\operatorname{GConv}_{5\times5}(\cdot)$ denotes $5\times5$ group convolutional operation.

To extract discriminative information from the optical flow, $\mathcal{F}_{mid}$ is passed through a $1\times1$ convolutional module to generate $\mathcal{F}^{\prime}_{mid}$. This feature map is then fed into the Flow module, yielding $\mathcal{F}_{f}$ and $\mathcal{F}^{\prime}_{f}$, respectively. The Flow module performs an operation that uniformly partitions the input $\mathcal{F}^{\prime}_{mid}$ into two tensor blocks, $\mathcal{F}_{f}$ and $\mathcal{F}^{\prime}_{f}$, along the channel dimension, returning a tuple of these tensor blocks. The details are presented as follow:
\begin{equation}
    \label{eqn:10}
    \mathcal{F}_{f},\mathcal{F}^{\prime}_{f} = \operatorname{Flow}(\operatorname{Conv}_{1\times1}(\mathcal{F}_{mid}))
\end{equation}

The $\mathcal{F}_{in}$ is fed into a $1\times1$ convolutional module to generate $\mathcal{F}_{m}$. Then, $\mathcal{F}_m$ is input into the Multiscale Aware Integration (MSAI) module to yield $\mathcal{F}_{b}$. $\mathcal{F}_{b}$ and $\mathcal{F}_{f}$ are input into the Warp module to obtain $\mathcal{F}^{\prime}_{Branch}$. Similarly, $\mathcal{F}_{Branch}$ is derived from $\mathcal{F}^{\prime}_{b}$ and $\mathcal{F}^{\prime}_{f}$. The $\operatorname{Warp}$ module initially constructs a regular grid that matches the dimensions of the original image, and then generates a deformation field from this grid to align the image. Using the deformation function, the original image and the flow field serve as parameters. The deformation field is incorporated into the regular grid, which is then used to resample the original image. This method enables mutual supervision of the aligned left and right images through consistency, effectively addressing misclassification issues due to the height displacement of buildings. We obtain $\mathcal{F}^{\prime}_{out}$ by subtracting $\mathcal{F}^{\prime}_{Branch}$ from $\mathcal{F}^{\prime}_{b}$, formulated as: 
\begin{equation}
    \label{eqn:11}
    \mathcal{F}^{\prime}_{out} = \operatorname{Warp}(\mathcal{F}_{in}, \mathcal{F}_{f}) - \mathcal{F}^{\prime}_{b}
\end{equation}

Similarly, $\mathcal{F}_{out}$ is obtained, and $\mathcal{F}_{out}$ and $\mathcal{F}^{\prime}_{out}$ are fused by concatenation operation. The structure of MSAI is described below.

Conventional optical flow-based change detection methods often suffer from restricted receptive fields in capturing multiscale building structures, particularly failing to resolve the intrinsic scale variance problem in urban environments where buildings exhibit diverse sizes and complex spatial arrangements. While recent adaptations of foundation models like SAM have employed Transformer-based decoders, their fixed-scale attention mechanisms are suboptimal for detecting subtle structural changes and precise boundary localization in high-resolution remote sensing images. The MSAI module introduces three key innovations: 1) a hierarchical dilation architecture that systematically addresses scale variance through growing receptive fields; 2) a learnable spatial weighting mechanism that dynamically prioritizes building edges over homogeneous regions; and 3) an ensemble learning strategy that synergizes multi-granularity features beyond standard residual connections. As shown in Figure \ref{fig:MSAFA} (b), $\mathcal{F}_{m}$ passes through multiple $3 \times 3$ dilated convolution modules with different dilation rates of 5, 7, 9, 11. These operations produce $\mathcal{F}_{mi}$, where $i \in \{5,7,9,11\}$. These four feature maps are fed into $3\times3$ group convolution and point convolution, respectively. These operations refine the edge details of the feature maps. Finally, the feature maps are summed to generate $\mathcal{F}_{dilation}$, as given by the following equation:

\begin{equation}
    \label{eqn:12}
    \begin{aligned}
    &\mathcal{F}_{mi}= \operatorname{DConv_{3 \times 3}}(\mathcal{F}_{m}), i=5,7,9,11\\
    &\mathcal{F}_{dilation}= \sum\operatorname{PConv}(\operatorname{GConv_{3 \times 3}}(\mathcal{F}_{mi})), i=5,7,9,11
    \end{aligned}
\end{equation}
in which $\operatorname{PConv}(\cdot)$ represents $1\times1$ convolution, $\operatorname{DConv_{3 \times 3}}(\cdot)$ indicates $3\times3$ dilation convolution, $\operatorname{GConv_{3 \times 3}}(\cdot)$ refers to $3\times3$ group convolution. 

Next, $\mathcal{F}_{dilation}$ is passed through the unfold module to capture the pixel-level information from local regions, enhancing the model's ability to detect fine details, such as edges and corners, and improving edge detection clarity. This step generates $\mathcal{F}_{u}$. $\mathcal{F}_{u}$ then enters `Get Weight' module to get $\mathcal{F}_{w}$. The reorganized weight structure allows the model to assign varying response strengths to features at each convolutional kernel position, enhancing its ability to capture building contours and details, particularly in complex regions. The weight $\mathcal{F}_{w}$ is then fed into the `Adjust Shape' module to generate $\mathcal{F}^{\prime}_{w}$, enabling the model to adjust the weight distribution based on features from different positions. 

The $\mathcal{F}_u$ undergoes two shape adjustment operations to align with the weights $\mathcal{F}_w$, allowing more fine-grained processing of building edge information. This improves the model's ability to capture subtle changes, especially in boundary details and building contours. This operation yields $\mathcal{F}^{\prime}_{u}$. The $\mathcal{F}^{\prime}_{u}$ and $\mathcal{F}^{\prime}_{w}$ are multiplied element-wise to produce $\mathcal{F}_{uw}$, effectively combining the reorganized receptive field features with the weights. The detailed procedure is presented below:

\begin{equation}
    \label{eqn:13}
    \begin{aligned}
    &\mathcal{F}_{u}= \operatorname{unfold}(\mathcal{F}_{dilation})\\
    &\mathcal{F}^{\prime}_{w}= \operatorname{Rearrange}(\operatorname{GW}(\mathcal{F}_{u}))\\
    &\mathcal{F}^{\prime}_{u}= \operatorname{Rearrange}(\operatorname{Rearrange}(\mathcal{F}_{u}))\\
    &\mathcal{F}_{uw}= \mathcal{F}^{\prime}_{u} \cdot \mathcal{F}^{\prime}_{w}
    \end{aligned}
\end{equation}
in which $\operatorname{unfold}(\cdot)$ is the operation of unfold, $\operatorname{Rearrange}(\cdot)$ represents the operation of `Adjust Shape', $\operatorname{GW}(\cdot)$ represents the operation of `Get Weight', and $\cdot$ indicates dot-Matrix operation.

The $\mathcal{F}_{uw}$ is processed through a $3\times3$ convolutional module and a $1\times1$ residual module. The feature map is then projected into ensemble learning module, which enhances feature extraction capabilities by integrating multiple convolutional modules. This module consists of three parallel $3\times3$ convolutional layers, with each module processing the input feature map independently to learn different feature representations, such as edges, textures, and local details. During the forward propagation, the feature map is passed through each convolutional layer, generating multiple feature maps. The feature maps are then stacked and averaged to combine the outputs from all layers. The formula is presented as follows:

\begin{equation}
    \label{eqn:15}
    \begin{aligned}
    &\mathcal{F}^{\prime}_{uw}= \operatorname{Conv_{3\times3}}(\mathcal{F}_{uw})\\
    &\mathcal{F}_{r}= \operatorname{Conv_{1\times1}}(\mathcal{F}^{\prime}_{uw}) + \mathcal{F}^{\prime}_{uw}\\
    &\mathcal{F}_{sum}= \sum_{i}(\operatorname{Conv}_{3\times3}(\mathcal{F}_{r})), i=3\\
    &\mathcal{F}_{b} = \operatorname{Mean}(\operatorname{Stack}(\mathcal{F}_{sum}))
    \end{aligned}
\end{equation}
in which $\operatorname{Stack}(\cdot)$ represents the operation of ``stack".

\section{Experiments and Analysis}
\subsection{Datasets and Experimental Setup}
\subsubsection{Datasets}
To evaluate the performance of FAEWNet, we conduct experiments using three widely used building change detection datasets: LEVIR-CD (\citeauthor{Chen2020}, \citeyear{Chen2020}), S2Looking (\citeauthor{SL2021}, \citeyear{SL2021}) and WHU-CD (\citeauthor{JI2019}, \citeyear{JI2019}) datasets. 
\begin{itemize}
\item The \textbf{LEVIR-CD} dataset comprises of 637 pairs of very high-resolution images (0.5 m/pixel) from various cities in Texas, USA. Each image is $1024 \times 1024$ pixels in size. These bi-temporal images, captured over a period of 5 to 14 years, illustrate land-use changes, particularly construction growth.

\item The \textbf{S2Looking} dataset includes 5000 bi-temporal image pairs from rural regions worldwide with over 65,920 instances of changed buildings. Each image is $1024 \times 1024$ pixels in size, with a time interval of one-to-three years between the pairs. This dataset showcases a range of scene types such as villages, villas, industrial areas, and etc. 

\item The \textbf{WHU-CD} dataset consists of a pair of high-resolution images with the size of $32,507\times15,354$. These images document 12,796 buildings in 2012 and 16,077 buildings in 2016. Following \cite{bandara2022Revisiting}, the original images are cropped into non-overlapping image patches of
size $256\times256$.

\end{itemize}
\subsubsection{Evaluation Metrics}
For quantitative evaluation, we utilize four evaluation metrics: intersection over union ($IoU$),  precision ($Pr$), recall ($Rc$), and F1-score ($F1$).

\subsubsection{Implementation Details}
Our experiments are conducted on an NVIDIA GeForce GTX4090 GPU with Python 3.8.10 and are implemented in Pytorch. We utilize the AdamW optimizer for all models, with an initial learning rate of 0.0001 for the LEVIR-CD, S2Looking, and WHU-CD datasets. Specifically, for the Changer (ResNet18) and IDA-SiamNet (ResNet18) models, the learning rate is set to 0.005 for the LEVIR-CD dataset, and 0.001 for the S2Looking and WHU-CD datasets, as described in \cite{Li2024}. For the TTP, the learning rate is set to 0.0004 for the LEVIR-CD and S2Looking datasets. For FAEWNet model, the learning rate is set to 0.0002 for the LEVIR-CD dataset and 0.00025 for the S2Looking dataset. 

Due to limited GPU memory, the batch size for TTP and FAEWNet are set to 2, while other models utilize a batch size of 8. The data type for the FAEWNet is set to ``float16" to reduce memory usage. All models are trained for 719 epochs (it is equal to 40,000 iterations while batchsize is set to 8), 80,000 iterations and 100 epochs for the LEVIR-CD, the S2Looking and the WHU-CD datasets, respectively. Data augmentation techniques, including random cropping (for LEVIR-CD and S2Looking, images are divided into patches of $512\times512$ pixels, while the WHU-CD dataset retains $256\times256$ pixels without additional cropping), flipping, and photometric distortion, are applied to the training dataset. Additionally, the order of bi-temporal images is randomly shuffled to increase training diversity.

\subsection{Comparison with State-of-the-Art Methods}
To evaluate the effectiveness of our proposed method, we compare FAEWNet with thirteen existing change detection methods.
\begin{table*}[!htb]
\caption{Quantitative measures for all comparative methods on \textit{LEVIR-CD} dataset.}
\label{table:1}
\centering
\small
\scalebox{0.78}{
\begin{tabular}{l c c c c c c c c c}
\toprule

Methods ~ &Backbone  &Pr(\%) &Rc(\%)  &F1(\%) & IoU(\%)\\
\midrule
TINYCD (\citeauthor{Codegoni2022}, \citeyear{Codegoni2022})& EfficientNet &88.87  &84.92  &86.85  &76.76 \\

SNUNet/16 (\citeauthor{Fang2022}, \citeyear{Fang2022})& - &91.89  &88.62   &90.23  &82.19 \\

DSIFN (\citeauthor{Zhang2020b}, \citeyear{Zhang2020b})& VGG16 &90.95       &90.42       &90.68       &82.95 \\

ChangerAD (\citeauthor{Fang2023}, \citeyear{Fang2023})& ResNet18 &93.54  &89.66   &91.56   &84.43 \\

Changer (\citeauthor{Fang2023}, \citeyear{Fang2023})& ResNet18 &93.34  &90.09   &91.68   &84.65 \\

IDA-SiamNet (\citeauthor{Li2024}, \citeyear{Li2024}) & ResNet18 &93.17  &90.54   &91.84   &84.91\\
\midrule
BiT (\citeauthor{Chen2021a}, \citeyear{Chen2021a})&  ResNet18+Transformer &92.26    &89.47   & 90.84  &83.22\\

ChangeFormer (\citeauthor{Bandara2022}, \citeyear{Bandara2022})& MiT-b1 &93.73     &90.15   &91.91   &85.03 \\

ChangerAD (\citeauthor{Fang2023}, \citeyear{Fang2023})& MiT-b1 &93.67  &90.04   &91.82   &84.88  \\

Changer (\citeauthor{Fang2023}, \citeyear{Fang2023})& MiT-b1 &93.57  &90.05   &91.78   &84.80 \\

IDA-SiamNet (\citeauthor{Li2024}, \citeyear{Li2024}) & MiT-b1 &93.49  &90.74   &92.10   &85.35  \\
\midrule
BAN (\citeauthor{LiKaiyu2024}, \citeyear{LiKaiyu2024}) & VFM  &93.58 &89.88  &91.70   &84.67\\ 

TTP (\citeauthor{Chen2023}, \citeyear{Chen2023}) & VFM  &\textbf{93.75} &90.99  &92.35   &85.79\\ 
\rowcolor{blue!8}

FAEWNet & VFM &93.56 &\textbf{91.29} & \textbf{92.41}  &\textbf{85.89} \\

\bottomrule
\end{tabular}}
\end{table*}

\begin{figure*}[!htb]
\centering
\includegraphics[width=0.90\linewidth]{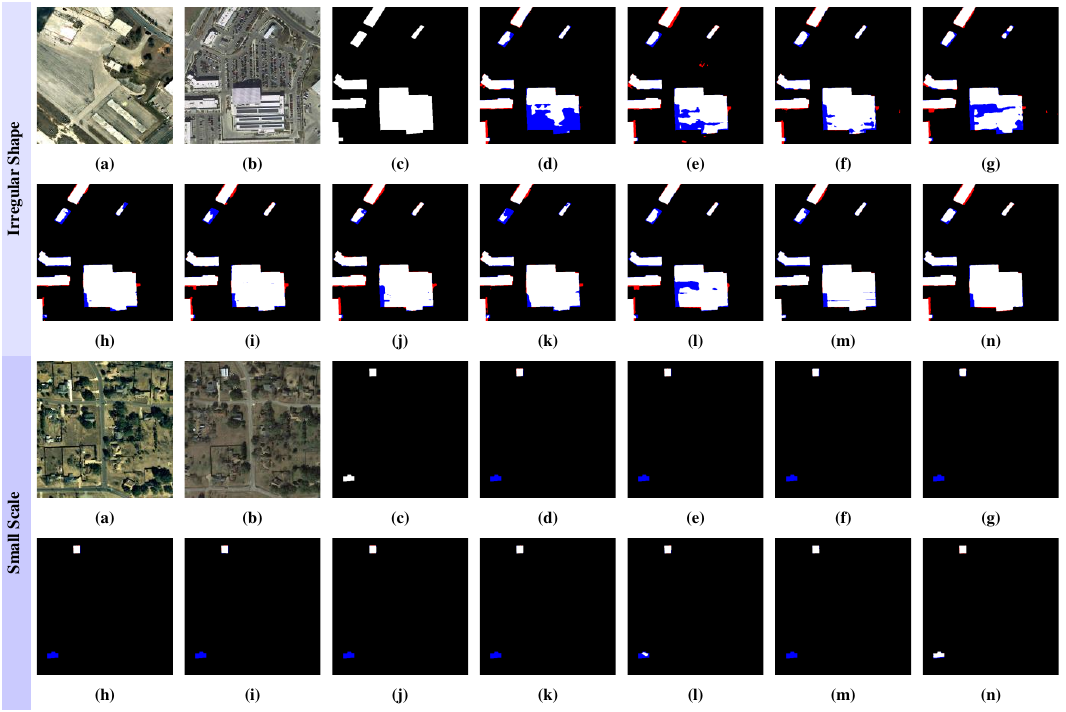}
\caption{Visual comparison of building change results on the LEVIR-CD dataset. (a) T1 image, (b) T2 image, (c) Change label, (d) ChangerAD (ResNet18), (e) Changer (ResNet18), (f) IDA-SiamNet (ResNet18), (g) BiT, (h) ChangeFormer, (i) ChangerAD (MiT-b1), (j) Changer (MiT-b1), (k) IDA-SiamNet (MiT-b1), (l) BAN, (m) TTP and (n) FAEWNet. White represents a true positive, black is a true negative, red indicates a false positive, and blue represents a false negative.}
\label{fig:x}
\end{figure*}

\subsubsection{Experiments on LEVIR-CD Dataset}
Table \ref{table:1} summarizes the performance of different methods on the LEVIR-CD dataset. FAEWNet consistently outperforms other approaches across the majority of evaluation metrics. It achieves a 0.30\% improvement in $Rc$, a 0.06\% increase in $F1$, and a 0.10\% enhancement in $IoU$ compared to the second-best method. These results underscore the robustness and effectiveness of FAEWNet in buidling change detection.

Figure \ref{fig:x} illustrates qualitative comparisons of building change detection results on the LEVIR-CD dataset under different scenarios. These visual results emphasize that FAEWNet excels in detecting building changes in complex scenes, particularly for buildings with irregular shapes, compared to other networks.
\begin{table*}[!htb]
\caption{Quantitative measures for all comparative methods on \textit{S2Looking} dataset.}
\label{table:2}
\centering
\small
\scalebox{0.78}{
\begin{tabular}{l c c c c c c c c c}
\toprule

Methods ~ & Backbone &Pr(\%) &Rc(\%)  &F1(\%)  & IoU(\%)\\
\midrule
TINYCD (\citeauthor{Codegoni2022}, \citeyear{Codegoni2022}) & EfficientNet    &69.33   &43.21   &53.24  &36.28\\

SNUNet/16 (\citeauthor{Fang2022}, \citeyear{Fang2022}) & -  &63.58  &52.32   &57.40  &40.25\\

DSIFN (\citeauthor{Zhang2020b}, \citeyear{Zhang2020b}) & VGG16   &52.44       &56.11       &54.21       &37.19  \\

ChangerAD  (\citeauthor{Fang2023}, \citeyear{Fang2023}) & ResNet18 &73.12  &59.22   &65.44   &48.63 \\

Changer (\citeauthor{Fang2023}, \citeyear{Fang2023}) & ResNet18 &72.97  &59.53   &65.57   &48.78 \\

IDA-SiamNet (\citeauthor{Li2024}, \citeyear{Li2024}) & ResNet18 &72.15  &60.55   &65.84   &49.08 \\
\midrule
BiT (\citeauthor{Chen2021a}, \citeyear{Chen2021a}) & ResNet18+Transformer &73.45    &55.95   &63.51   &46.53\\

ChangeFormer (\citeauthor{Bandara2022}, \citeyear{Bandara2022}) & MiT-b1 &72.34     &61.99   & 66.76  &50.11 \\

ChangerAD  (\citeauthor{Fang2023}, \citeyear{Fang2023}) & MiT-b1 &70.29  &63.32   &66.63   &49.96 \\

Changer (\citeauthor{Fang2023}, \citeyear{Fang2023}) & MiT-b1 &72.18     &62.16     &66.80    &50.15 \\

IDA-SiamNet (\citeauthor{Li2024}, \citeyear{Li2024}) & MiT-b1  &70.96  &63.87   &67.23   &50.64 \\
\midrule
BAN (\citeauthor{LiKaiyu2024}, \citeyear{LiKaiyu2024}) & VFM   &70.65 &62.93   &66.57   &49.89\\

TTP (\citeauthor{Chen2023}, \citeyear{Chen2023}) & VFM   &\textbf{74.81} &61.59  &67.56   &51.01 \\

\rowcolor{blue!8}
FAEWNet (Ours) & VFM  &73.08  &\textbf{64.68}   &\textbf{68.63}   &\textbf{52.24} \\

\bottomrule
\end{tabular}}
\end{table*}
\begin{figure*}[!htb]
\centering
\includegraphics[width=0.90\linewidth]{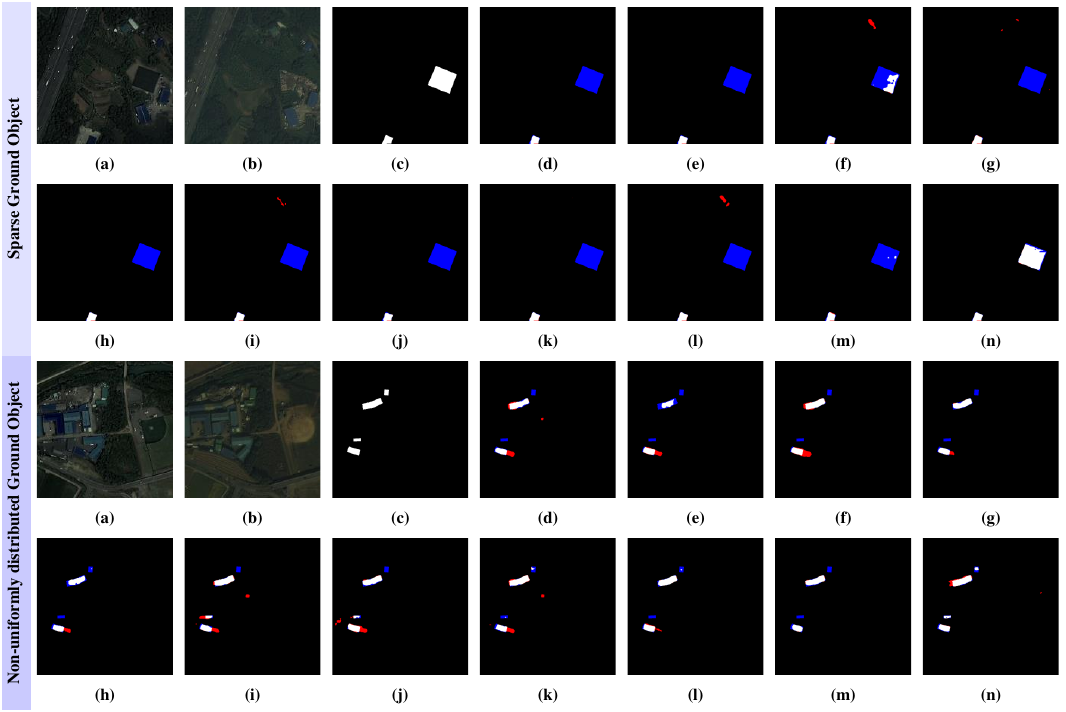}
\caption{Visual comparison of building change results on the S2Looking dataset. (a) T1 image, (b) T2 image, (c) Change label, (d) ChangerAD (ResNet18), (e) Changer (ResNet18), (f) IDA-SiamNet (ResNet18), (g) BiT, (h) ChangeFormer, (i) ChangerAD (MiT-b1), (j) Changer (MiT-b1), (k) IDA-SiamNet (MiT-b1), (l) BAN, (m) TTP and (n) FAEWNet. White represents a true positive, black is a true negative, red indicates a false positive, and blue stands as a false negative.}
\label{fig:y}
\end{figure*}
\subsubsection{Experiments on S2Looking Dataset}
Table \ref{table:2} presents the quantitative evaluation results for the S2Looking dataset. FAEWNet outperforms other methods, achieving a 3.09\% improvement in $Rc$, a 1.07\% increase in $F1$, and a 1.23\% enhancement in $IoU$ compared to the second-best method. While FAEWNet achieves the highest scores in $Rc$, $F1$, and $IoU$, its precision metric is slightly lower due to the presence of false positives in the detection results.

Figure \ref{fig:y} visualizes the building change detection results for the S2Looking dataset in challenging scenarios. Unlike the LEVIR-CD dataset, S2Looking images are characterized by significantly lower contrast, making change detection more challenging. FAEWNet provides clear delineations of building shapes and positions, while competing methods depict the outlines and edges of the changed buildings with less clarity. These results confirm FAEWNet's enhanced ability to detect building changes accurately, even under low-contrast conditions, outperforming other approaches.

\subsubsection{Experiments on WHU-CD Dataset}
Table \ref{table:3} presents quantitative evaluation results for various methods on the WHU-CD dataset. FAEWNet still outperforms the other methods, achieving a 0.58\% improvement in $Rc$, a 0.43\% enhancement in $F1$, and a 0.77\% increase in $IoU$ compared to the second-best approach. Notably, FAEWNet significantly outperforms other CNN-based and Transformer-based methods across the $Rc$, $F1$, and $IoU$ metrics.  

\begin{table*}[!htb]
\caption{Quantitative measures for all comparative methods on \textit{WHU-CD} dataset.}
\label{table:3}
\centering
\small
\scalebox{0.78}{
\begin{tabular}{l c c c c c c c c}
\toprule

Methods ~ &Backbone  &Pr (\%)&Rc(\%)  &F1(\%)  & IoU(\%)\\
\midrule
TINYCD (\citeauthor{Codegoni2022}, \citeyear{Codegoni2022}) & EfficientNet  &92.05   &85.00   &88.39  &79.19\\

SNUNet/16 (\citeauthor{Fang2022}, \citeyear{Fang2022}) & -  &91.76 &89.22   &90.47  &82.61\\

DSIFN (\citeauthor{Zhang2020b}, \citeyear{Zhang2020b})  & VGG16 &94.51      &89.75       &92.07       &85.30 \\

ChangerAD (\citeauthor{Fang2023}, \citeyear{Fang2023}) & ResNet18  &\textbf{97.26}  &88.68   &92.77   &86.52\\

Changer (\citeauthor{Fang2023}, \citeyear{Fang2023}) & ResNet18  &94.63  &90.44   &92.49   &86.03 \\

IDA-SiamNet (\citeauthor{Li2024}, \citeyear{Li2024}) & ResNet18  &96.29  &91.16   &93.66   &88.07 \\
\midrule
BiT (\citeauthor{Chen2021a}, \citeyear{Chen2021a})  & ResNet18+Transformer &94.97  &92.52   &93.73   &88.20 \\

ChangeFormer (\citeauthor{Bandara2022}, \citeyear{Bandara2022})& MiT-b1 &95.95  &92.91   &94.41   &89.40 \\

ChangerAD (\citeauthor{Fang2023}, \citeyear{Fang2023})& MiT-b1 & 96.24 & 92.67  & 94.42  & 89.43\\

Changer (\citeauthor{Fang2023}, \citeyear{Fang2023}) & MiT-b1 &95.26  &92.56   &93.89   &88.48 \\

IDA-SiamNet (\citeauthor{Li2024}, \citeyear{Li2024}) & MiT-b1 &95.83  &93.18   &94.49   &89.55\\
\midrule
BAN (\citeauthor{LiKaiyu2024}, \citeyear{LiKaiyu2024}) & VFM   &95.84 &93.18   &94.49  &89.56 \\

TTP (\citeauthor{Chen2023}, \citeyear{Chen2023}) & VFM   &95.51 &93.62   &94.56   &89.68 \\
\rowcolor{blue!8}
FAEWNet (Ours) & VFM  &95.79  &\textbf{94.20}   &\textbf{94.99}   &\textbf{90.45}\\
\bottomrule
\end{tabular}}
\end{table*}

\begin{figure*}[!htb]
\centering
\includegraphics[width=0.90\linewidth]{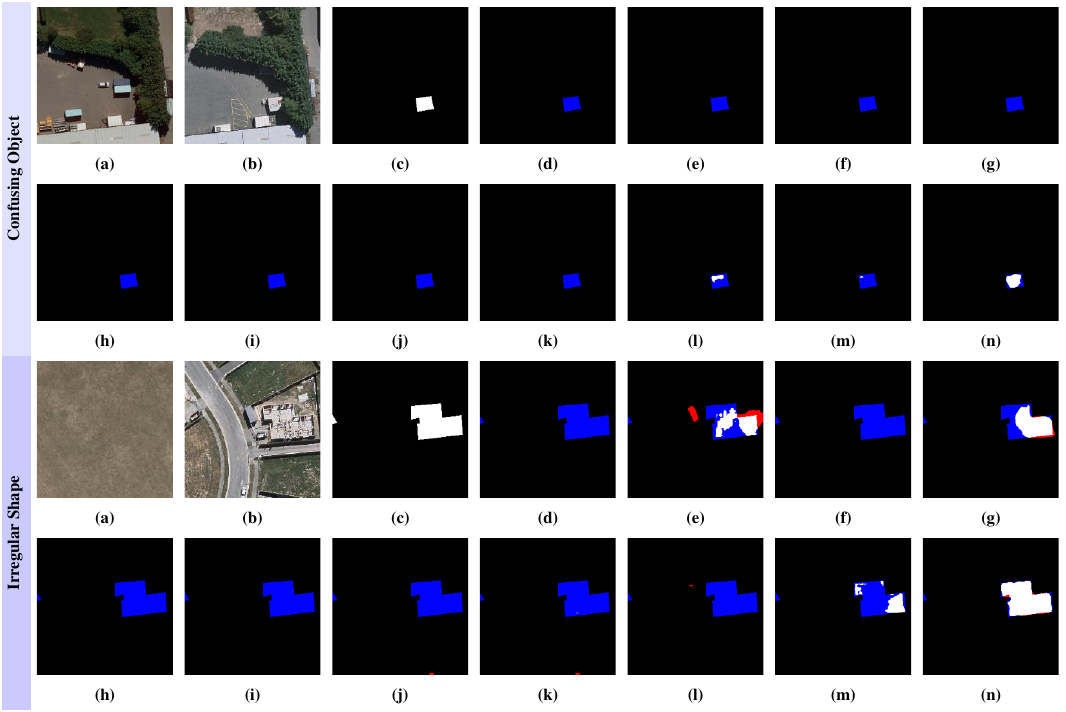}
\caption{Visual comparison of building change results on the WHU-CD dataset. (a) T1 image, (b) T2 image, (c) Change label, (d) ChangerAD (ResNet18), (e) Changer (ResNet18), (f) IDA-SiamNet (ResNet18), (g) BiT, (h) ChangeFormer, (i) ChangerAD (MiT-b1), (j) Changer (MiT-b1), (k) IDA-SiamNet (MiT-b1), (l) BAN, (m) TTP and (n) FAEWNet. White represents a true positive, black is a true negative, red indicates a false positive, and blue represents a false negative.}
\label{fig:z}
\end{figure*}
 
The qualitative results in Figure \ref{fig:z} offer a visual comparison of building change detection networks on the WHU-CD dataset. FAEWNet can accurately detect confusing and irregular building changes, further confirming its robustness across diverse scenarios.

\begin{figure}[!htb]
\centering
\includegraphics[width=1\linewidth]{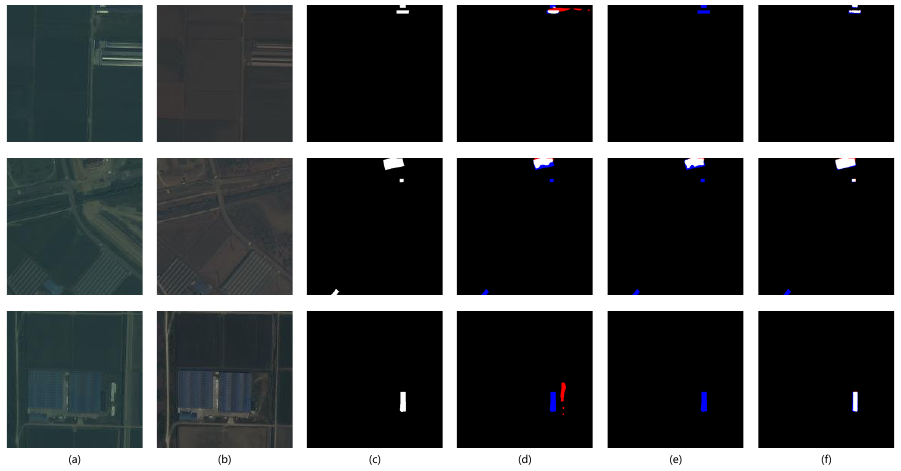}
\vspace{-0.4cm}
\caption{Visualization Performance with Imbalanced Building Distribution. (a) T1 image, (b) T2 image, (c) Change label, (d) BAN (\citeauthor{LiKaiyu2024}, \citeyear{LiKaiyu2024}), (e) TTP (\citeauthor{Chen2023}, \citeyear{Chen2023}), (f) FAEWNet. White represents a true positive, black is a true negative, red indicates a false positive, and blue stands as a false negative.}
\label{fig:shili2}
\end{figure}
\subsection{Ablation Study}
\subsubsection{Validation of the impact of each module}
As shown in Figure \ref{fig:shili2}, DAFA addresses the issue of imbalanced distribution. We further conduct a series of ablation experiments to validate the effectiveness of the DAFA and MSAFA modules. The baseline consists of a LoRA-based encoder and a decoder composed of simple FPN, a projection layer. All the experiments are implemented with the same training configuration. 

\begin{table}[!htbp]
	\centering
	\caption{Ablation experiments for the DAFA and MSAFA modules on \textit{S2Looking} dataset.}
	\label{table:5}
	\renewcommand{\arraystretch}{1.2}
	\setlength{\tabcolsep}{5pt}
	\small
	{
		\begin{tabular}{ccccc}
            \toprule
			\textbf{+ DAFA} & \textbf{+ MSAFA}&\textbf{Rc(\%)}&\textbf{F1(\%)} & \textbf{IoU(\%)} 
			\\ \midrule

			  \ding{55} & \ding{55} &62.10     &67.66  &51.12
			\\ 
			  \ding{51} & \ding{55} &62.77  &68.46   &52.05   
			\\ 
			  \ding{55} & \ding{51} &64.31   &68.41  &51.99 
			\\ 
			  \ding{51} & \ding{51} &\textbf{64.68}    &\textbf{68.63}   &\textbf{52.24}
			\\ \bottomrule
			
		\end{tabular}
	}
\end{table}
Experiments with various module configurations are conducted on S2Looking dataset to evaluate the impact of each module on the building change detection task. The results are presented in Table \ref{table:5}, utilizing three metrics: $Rc$, $F1$, and $IoU$. These findings indicate that integrating either of the two modules enhances the baseline performance, with the best results obtained when all modules are utilized together. This confirms the effectiveness of the proposed DAFA, and MSAFA modules. 
\begin{figure}[!htb]
\centering
\includegraphics[width=1\linewidth]{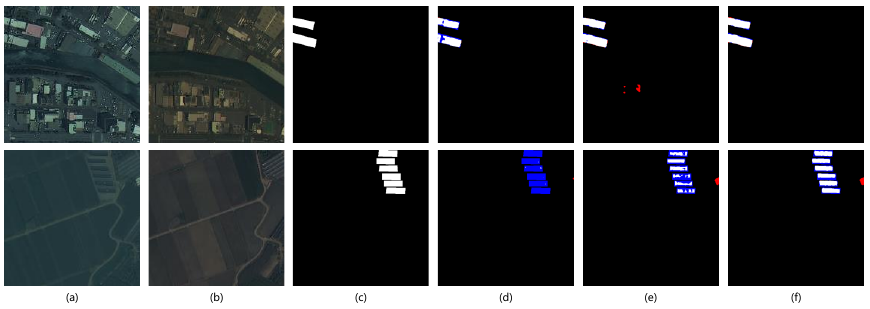}
\vspace{-0.4cm}
\caption{Visualization results of the ablation experiments. (a) T1 image, (b) T2 image, (c) Change label, (d) Baseline, (e) Baseline + DAFA, (f) FAEWNet. White represents a true positive, black is a true negative, red indicates a false positive, and blue stands as a false negative.}
\label{fig:ablation_fig}
\end{figure}
Furthermore, to visually assess the impact of these modules, we show the corresponding change detection results in Figure \ref{fig:ablation_fig}. These results demonstrate that the proposed method improves significantly with the addition of these modules, leading to the identification of more changed areas and fewer false alarms. Therefore, the proposed FAEWNet achieves the best performance in building change detection.

\subsubsection{Analysis of the Effectiveness of DAFA}
We analyze the impact of different components of DAFA on building change detection performance, as shown in Table \ref{table:10}. In DAFA, ``w/o DACA" indicates the DACA module is not used. ``Only one Linear" represents that there is one Linear instead of three parallel Linear modules. Comparing the results in the first and last rows of the table, we can conclude that the DACA module collects various channel statistics to mitigate information loss. This process captures the distribution of buildings. A comparison of the results in the second and fourth rows of the table shows that DFT is effective in filtering out high-frequency noise or other interferences while preserving and enhancing frequency components represent true building changes. The comparison between the third and last rows shows that three parallel Linear modules allow the model to learn small-scale representations and the distribution of buildings, further improving performance.
\begin{table}[!htb]
\caption{Ablation experiments for different parts of DAFA on \textit{S2Looking} dataset.}
\label{table:10}
\centering
\small
\begin{tabular}{l c c c c}
\toprule

Methods ~   &Rc(\%)  &F1(\%)  & IoU(\%) \\
\midrule
w/o DACA   &61.25  &67.34  &50.76    \\

w/o DFT    &63.22   &68.41  &51.99    \\

Only one Linear   &62.47     &68.51      &52.10 \\

All   &\textbf{64.68}       &\textbf{68.63}      &\textbf{52.24}\\
\bottomrule
\end{tabular}
\end{table}

\subsubsection{Analysis of the Effectiveness of MSAI}
\begin{figure}[!htp]
\centering
{\includegraphics[width=0.5\linewidth]{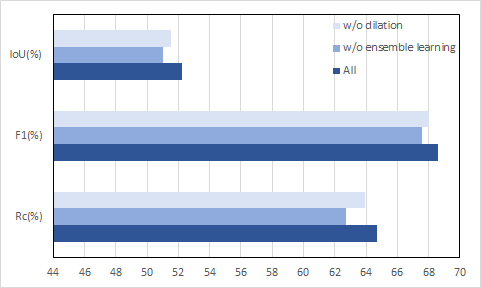}}
\caption{Ablation experiments for different parts of MSAI on \textit{S2Looking} dataset.}
\label{fig:AblationexperimentsfordifferentpartsofMSAIonS2Lookingdataset}
\end{figure}

To further illustrate the advantages of dense dilation convolutional module and ensemble learning in MSAI, we do the following ablation experiments. In the MSAI module, ``w/o dilation" indicates that dense dilation convolution, group convolution, and point convolution are not utilized. As shown in Figure \ref{fig:AblationexperimentsfordifferentpartsofMSAIonS2Lookingdataset}, MSAI yields the highest performance in terms of $Rc$, $F1$, and $IoU$. The ``w/o dilation" and ``All" illustrate that dense dilation convolution and group convolution help the network refine the edge information in the feature map. A comparison of the ``w/o ensemble learning" and ``All" shows that ensemble learning allows the model to capture more comprehensive information from the input data, improving the detection of edges, details, and local changes in the buildings.

\subsection{Discussion}
\subsubsection{Comparison between Different Adapter Modules}
\begin{figure}[!htp]
\centering
{\includegraphics[width=0.5\linewidth]{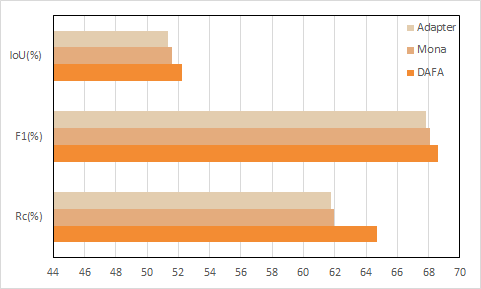}}
\caption{Comparison experiments for different Adapters on \textit{S2Looking} dataset.}
\label{fig:AblationexperimentsfordifferentAdaptersonS2Lookingdataset}
\end{figure}

We evaluate the performance of various Adapter modules on building change detection tasks. The Adapter module proposed by \cite{houlsby2019} adopts a lightweight bottleneck architecture. The Mona Adapter (\citeauthor{yin2024}, \citeyear{yin2024}) incorporates vision-specific filters to enhance its processing of visual features. As shown in Figure \ref{fig:AblationexperimentsfordifferentAdaptersonS2Lookingdataset}, the DAFA achieves the highest performance across all metrics, indicating its ability to precisely extract building boundaries. Compared to other Adapter modules, DAFA demonstrates a remarkable advantage in addressing the domain gap between natural and remote sensing images, further validating its effectiveness in building change detection applications.
\subsubsection{Comparison between Different Optical Flow Methods}

To further demonstrate the advantages of MSAFA, we compare it with prior interpreter (PI) (\citeauthor{LMX2022}, \citeyear{LMX2022}) and dynamic-deformable dual-alignment fusion (D$^3$AF) (\citeauthor{Li2024}, \citeyear{Li2024}). As shown in Table \ref{table:9}, MSAFA achieves the highest performance in $Rc$, $F1$, and $IoU$. The MSAFA module excels at capturing multiscale contextual information to address registration errors. In comparison with PI and D$^3$AF, our module enhances the network's ability to perceive building changes from multiple orientations, further improving detection accuracy.
\begin{table}[!htb]
\caption{Comparison experiments for the different optical flow methods on \textit{S2Looking} dataset.}
\label{table:9}
\centering
\small
\begin{tabular}{l c c c c c}
\toprule

Methods ~   &Rc(\%)  &F1(\%)  & IoU(\%) \\
\midrule
PI (\citeauthor{LMX2022}, \citeyear{LMX2022}) & 62.08& 67.25& 50.65\\

D$^3$AF (\citeauthor{Li2024}, \citeyear{Li2024}) &61.83  &67.87  &51.36    \\

MSAFA   &\textbf{64.68}      &\textbf{68.63}       &\textbf{52.24} \\

\bottomrule
\end{tabular}
\end{table}

\subsubsection{Analysis of Parameters in DFT}

We conduct comparison experiments to justify the decision of using only the real part of the DFT. As shown in Table \ref{table:12}, using only the real part of the DFT yields the best performance in terms of $Rc$, $F1$, and $IoU$. The real part contains the most significant spatial pattern and structural information of buildings. In contrast, the imaginary part is typically associated with phase information, which is more sensitive to noise and small fluctuations, and is less relevant when focusing on large-scale building changes. By retaining only the real part, we effectively discard unnecessary information and focus on the more stable part of the buildings, which is key to accurately detecting building changes.
\begin{table}[!htb]
\caption{Sensitivity comparison and analysis of the parameters of DFT on the \textit{S2Looking} dataset.}
\label{table:12}
\centering
\small
\begin{tabular}{l c c c c c c}
\toprule

Methods ~   &Rc(\%)  &F1(\%)  & IoU(\%) \\
\midrule
w/o DFT    &63.22 &68.41  &51.99   \\

Amplitude &61.07  &67.80  &51.29    \\

DFT (Imag) &61.74  &67.80  &51.29    \\ 

DFT (Real)   &\textbf{64.68}       &\textbf{68.63}      &\textbf{52.24}\\

\bottomrule
\end{tabular}
\end{table}

\subsubsection{Analysis of DAFA Adapter in Different Positions}

\begin{figure}[!htp]
\centering
{\includegraphics[width=0.5\linewidth]{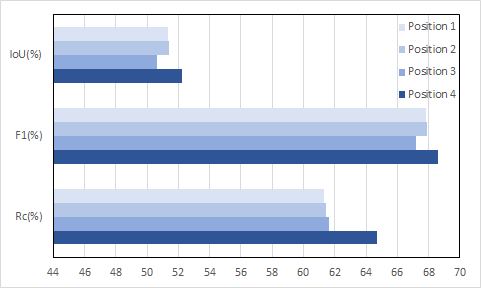}}
\caption{Sensitivity comparison and analysis of the position of DAFA on the \textit{S2Looking} dataset.}
\label{fig:SensitivitycomparisonandanalysisofthepositionofDAFAontheS2Lookingdataset}
\end{figure}

To demonstrate the effectiveness of DAFA, we present some comparison experiments analyzing its positioning. The DAFA module fine-tunes SAM's different global attention layers. We examine a ViT set to ``Large" as an example. As shown in Figure \ref{fig:SensitivitycomparisonandanalysisofthepositionofDAFAontheS2Lookingdataset}, ``Position 4" yields the best performance in terms of $Rc$, $F1$, and $IoU$. Experimental results demonstrate that placing the DAFA at the final global attention layer achieves the optimal performance. This design effectively integrates global information, enhancing sensitivity to building morphology, boundary changes, and distributions. Additionally, the DAFA improves the flexibility and refinement of feature representation, facilitating the detection of subtle changes, thereby significantly improving the overall performance of change detection.

\subsubsection{Parameter and MAC Analysis of foundation models}
\begin{table}[!htb]
\caption{Parameters and MACs on \textit{LEVIR-CD}, \textit{S2Looking} and \textit{WHU-CD} dataset.}
\label{table:6}
\centering
\small
\begin{tabular}{l c c c c}
\toprule

Methods ~   &Parameters (G)  &MACs (T) \\
\midrule
BAN (\citeauthor{LiKaiyu2024}, \citeyear{LiKaiyu2024})  &0.232/0.232/0.232  &0.330/0.330/0.298 \\

TTP (\citeauthor{Chen2023}, \citeyear{Chen2023})  &0.320/0.320/0.319  &0.839/0.839/0.275 \\

FAEWNet   &0.364/0.364/0.363       &0.922/0.922/0.296 \\

\bottomrule
\end{tabular}
\end{table}

The Parameters (Params) and multiply-accumulate operations (MACs) for the foundation models are shown in Table \ref{table:6}.  Experimental results reveal that the number of Params and MACs significantly impacts model performance. In these three datasets, although FAEWNet requires slightly more computations than other methods, its overall performance is superior, and we plan to explore a lightweight version of FAEWNet in future works. 











\section{Conclusion}
In this paper, we propose a new FAEWNet which utilizes the SAM encoder to extract rich visual features from remote sensing images. To address the fact that existing adapter-based methods remain constrained by imbalanced building distribution and small-scale buildings, our FAEWNet incorporates a DAFA Adapter to aggregate task-specific changed information, enhancing the network's ability to extract building distribution and edge details. Furthermore, traditional alignment approaches that rely on optical flow estimation are insufficient for accurate building change detection, as they tend to be influenced by background noises. To address this limitation, we design a novel flow module that refines building edge extraction, enhances the perception of changed buildings and mitigates background noises. Experimental results on three public building change detection datasets substantiate that our method surpasses other state-of-the-art methods. 

However, our FAEWNet can be further improved, particularly in terms of computational efficiency and mitigating overfitting due to an excessive number of parameters. Future research will focus on exploring lightweight foundation models and their applications in broader multimodal remote sensing datasets. We also plan to integrate semi-supervised learning approaches to address scenarios with limited datasets. 

\section*{Acknowledgements}
This work was supported in part by the National Natural Science Foundation of China under Grant 62271418, and in part by the Natural Science Foundation of Sichuan Province under Grant 2023NSFSC0030 and 2025ZNSFSC1154, in part by the Postdoctoral Fellowship Program and China Postdoctoral Science Foundation under Grant Number BX20240291, and in part by the Fundamental Research Funds for the Central Universities under Grant 2682025CX033.



\bibliographystyle{elsarticle-harv}
\bibliography{mybibfile}
\end{document}